\documentclass[10pt,twocolumn,letterpaper]{article}

\usepackage{wacv}              % To produce the CAMERA-READY version
\usepackage{graphicx}
\usepackage{amsmath}
\usepackage{amssymb}
\usepackage{booktabs}
\usepackage{times}
\usepackage{soul}
\usepackage{url}
\usepackage[hidelinks]{hyperref}
\usepackage[utf8]{inputenc}
\usepackage{amsmath}
\usepackage{amsthm}
\usepackage{booktabs}
\usepackage{algorithm}
\usepackage{algorithmic}
\usepackage[switch]{lineno}
\usepackage{mathtools}
\usepackage{amsmath}
\usepackage{amssymb,amsfonts}
\usepackage{multirow}
\usepackage{multicol}
\usepackage{color}
\usepackage{soul}

\usepackage{enumitem}
\usepackage{booktabs}
\usepackage{siunitx}
\usepackage[table]{xcolor}
\usepackage{tabularx}
\usepackage{xcolor}
\usepackage{adjustbox}
\usepackage{parskip}
\usepackage{array}
\usepackage{booktabs}
\usepackage{multirow}
\usepackage{multicol}
\usepackage{tabularx}
\usepackage{graphicx}
\usepackage{booktabs}
\usepackage{xcolor} 

\usepackage[accsupp]{axessibility} % Improves PDF readability for those with disabilities.

\definecolor{mygreen}{RGB}{0,200,0}
\definecolor{myred}{RGB}{200,0,0}
\definecolor{mydeepblue}{RGB}{51,164,255}

\usepackage{pifont}
\newcommand{\cmark}{\ding{51}}%
\newcommand{\xmark}{\ding{55}}%

\DeclareMathOperator*{\argmin}{arg\,min}
\newcommand{\new}[1]{{\color{black}#1}}

\usepackage[capitalize]{cleveref}
\crefname{section}{Sec.}{Secs.}
\Crefname{section}{Section}{Sections}
\Crefname{table}{Table}{Tables}
\crefname{table}{Tab.}{Tabs.}

\def\wacvPaperID{1119} % *** Enter the WACV Paper ID here
\def\confName{WACV}
\def\confYear{2025}

\begin{document}

\title{OTCXR: Rethinking Self-supervised Alignment using Optimal Transport for Chest X-ray Analysis}
\author{
Vandan Gorade\textsuperscript{1,\textdagger}, 
Azad Singh\textsuperscript{2,\textdagger}, 
Deepak Mishra\textsuperscript{2}\\
\textsuperscript{1}Northwestern University\\
\textsuperscript{2}Indian Institute of Technology Jodhpur\\
{\tt\small vandan.gorade@northwestern.edu, singh.63@iitj.ac.in,  dmishra@iitj.ac.in}\\
\textsuperscript{\textdagger}These authors contributed equally to this work.
}

\maketitle
%%%% ijcai24.tex
% \footnotetext[0]{ Azad Singh and Vandan Goarde are joint first authors, contributed equally to this work.}
% \footnotetext[1]{Indian Institute of Technology Jodhpur, Jodhpur, RJ 342030 India}
% \footnotetext[2]{Northwestern University, Evanston, IL 60208, USA}

\begin{abstract}
Self-supervised learning (SSL) has emerged as a promising technique for analyzing medical modalities such as X-rays due to its ability to learn without annotations. However, conventional SSL methods face challenges in achieving semantic alignment and capturing subtle details, which limits their ability to accurately represent the underlying anatomical structures and pathological features. To address these limitations, we propose OTCXR, a novel SSL framework that leverages optimal transport (OT) to learn dense semantic invariance. By integrating OT with our innovative Cross-Viewpoint Semantics Infusion Module (CV-SIM), OTCXR enhances the model's ability to capture not only local spatial features but also global contextual dependencies across different viewpoints. This approach enriches the effectiveness of SSL in the context of chest radiographs. Furthermore, OTCXR incorporates variance and covariance regularizations within the OT framework to prioritize clinically relevant information while suppressing less informative features. This ensures that the learned representations are comprehensive and discriminative, particularly beneficial for tasks such as thoracic disease diagnosis. We validate OTCXR's efficacy through comprehensive experiments on three publicly available chest X-ray datasets. Our empirical results demonstrate the superiority of OTCXR over state-of-the-art methods across all evaluated tasks, confirming its capability to learn semantically rich representations.
% Open-source code will be made available after the peer-review process.
\end{abstract}

\section{Introduction}
Self-supervised learning (SSL) has gained significant momentum in deep learning in recent years~\cite{mocov2,simclr,barlow,byol,gorade2023large}. In the medical image analysis domain, where large-scale annotated datasets are challenging to obtain, SSL is particularly useful due to its ability to harness unlabeled data~\cite{medaug,mococxr,micle}. However, conventional SSL methods, such as SimCLR~\cite{simclr}, MoCo~\cite{mococxr}, BYOL~\cite{byol}, Barlow-Twins~\cite{barlow}, etc., though effective in various domains, encounter challenges when applying to chest X-rays(CXRs). The interpretation of CXRs necessitates a deep understanding of anatomical structures and pathological indicators for precise diagnosis and treatment decisions. This requirement poses a significant challenge for SSL methods that focus on generic feature learning while often discarding the clinically relevant patterns~\cite{multi-gcn,synergynet,semiGCN,micle,pacl}.

Conventional SSL methods treat the entire image as a collection of isolated features, thereby missing local information and global contextual dependencies between different anatomical regions. 
Moreover, local variations in CXRs are particularly useful because of the coexistence of anatomical heterogeneity and visual similarity of different regions. Heterogeneity refers to visual distinctions between instances of the same disease or anatomical structures~\cite{zheng2021ressl}. For instance, the shape and size of a nodule may vary based on its stage and location.
Conversely, visual similarity refers to the challenges in distinguishing between different diseases or structures that share common visual characteristics~\cite{xing2021categorical}. For instance, consolidation and pneumonia are labeled separately even though they are the types of lung opacities and have a natural relation. In the context of SSL pre-training for CXRs, it becomes essential for models to capture these local variations. Failure to do so may result in trivial solutions and hinder the acquisition of representative feature space, essential for meaningful CXR analysis~\cite{zhao2022cross,wang2022rcl}. 

Addressing these challenges, we propose rethinking the concept of alignment through invariance by introducing the concept of dense semantic invariance, potentially improving SSL pre-training in the context of CXRs. It empowers the model to capture precise anatomical and pathological attributes regardless of differences in viewpoints, orientations, and imaging conditions. To this end, we propose OTCXR, a novel objective that replaces existing invariance-based alignment strategies in SSL with dense semantic invariance, modeled using Optimal Transport Theory (OT)~\cite{luise2018differential}. Further, the OTCXR integrates the Cross-Viewpoint Semantics Infusion Module (CV-SIM) to enhance the model's ability to capture global contextual dependencies across diverse viewpoints. The OT solver utilizes the resulting CV-SIM representations as marginal constraints in the transport plan, minimizing the total cost of transporting features between images. This approach effectively improves the alignment of semantically relevant features. Furthermore, to ensure that the acquired representations are informative and less redundant, we apply variance and covariance regularizations~\cite{vicreg} during the representation learning process. The integration of variance and covariance regularization within the OT framework enhances the ability of the model to generate stable and semantically aligned representations. Our contributions are summarized as follows:

\begin{itemize}
    \item \new{We introduce OTCXR, a novel SSL framework that formulates dense semantic invariance as an OT problem, enabling fine-grained feature alignment effective for chest X-ray analysis during pre-training. }
    
    \item \new{We propose a Cross-Viewpoint Semantics Infusion Module (CV-SIM) that complements the OT approach by capturing global semantic relationships, enhancing the framework's ability to learn clinically relevant features across different views and transformations.}
    
    \item \new{Through comprehensive evaluations on diverse chest X-ray datasets (NIH-Chest X-ray14, Vinbig-CXR, and RSNA), we demonstrate OTCXR's effectiveness in various medical imaging tasks, particularly in limited data scenarios, showcasing consistent performance improvements over existing SSL methods.}
\end{itemize}

\begin{figure*}[ht]\centering
\def\svgwidth{\columnwidth}
\includegraphics[width=\linewidth,scale=1.0]{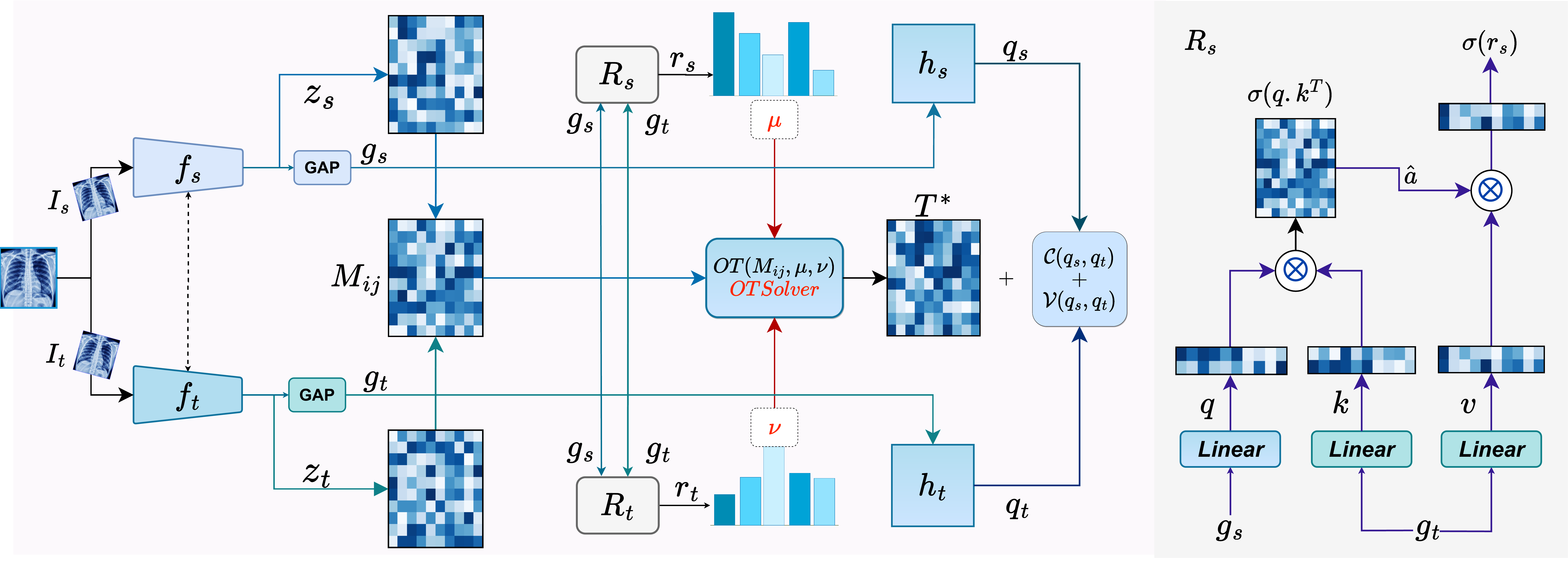}
\caption{Architecture of the OTCXR Framework. $I_s$ and $I_t$ are two augmented versions of $I$, pass through backbone encoders $f_s$ and $f_t$ to obtain $z_s$ and $z_t$ which OT solver subsequently utilizes along with the cost matrix $M$. $g_s$ and $g_t$ are the feature vectors after the global average pooling (GAP) layer, which are expanded to $q_s$ and $q_t$ to increase representational variability.}
\label{fig:workflow_1}
\end{figure*}

\section{Background and Related Works}
\textbf{Self-supervised Learning.}
 The current SSL approaches employ a joint embedding architecture to acquire representations invariant to diverse perspectives~\cite{simclr,byol,vicreg}. These techniques have varying ways of circumventing the collapse of solutions. Negative sampling is employed by contrastive methods~\cite{simclr,mocov2,mococxr,medaug} to push dissimilar samples apart from each other while clustering methods such as proposed in~\cite{SwaV} ensure an equitable distribution of samples within clusters. Non-contrastive techniques~\cite{byol,simclr,barlow,vicreg,dino}, which serve as the alternative of contrastive methods, maintain the representations' information content either by adopting architectural constraints such as asymmetric network, momentum encoder, stop-gradient, etc. or by applying specific regularization~\cite{barlow,vicreg}. Unlike the aforementioned global methods, local methods concentrate on learning a group of local features that describe smaller segments of an image. For instance, \new{the authors in ~\cite{pixelpro} addressed} the challenge of pixel-level self-supervised representation learning, crucial for tasks like object detection and semantic segmentation by leveraging pixel-level contrastive learning. Similarly, DenseCL~\cite{denseCL}, a novel dense contrastive learning paradigm, introduced pixel-level SSL using a dense projection head and dense contrastive loss. 

\textbf{SSL for CXR Analysis.}
SSL has become an increasingly popular approach for CXRs~\cite{shurrab2022self}. Most recent works have shown the effectiveness of SSL for tasks such as segmentation, classification, and localization~\cite{zhang2022dive,mococxr,medaug,huang2021lesion,10666966}. The adoption of SSL techniques has predominantly leaned towards \new{utilizing} contrastive learning approaches. However, in stark contrast, the number of approaches embracing non-contrastive SSL within this domain remains notably limited. For instance, \new{the authors in~\cite{cho2023chess}} introduced CheSS, a publicly accessible model pre-trained with 4.8-M chest X-ray images through contrastive learning on a large dataset. \new{Multi-Instance Contrastive Learning (MICLe), is another approach presented in~\cite{micle},  harnesses the multiple studies of the subject to create positive pairs. In another study, authors proposed a collaborative SSL framework, DiRA, that unites discriminative, restorative, and adversarial learning to learn fine-grained representations~\cite{dira}}. MedAug~\cite{medaug} and PaCL~\cite{pacl} are other works based upon the contrastive approach that incorporates insightful metadata for selective positive and negative pairs. PCRL~\cite{pcrlv1,pcrlv2}, another contrastive approach that enhances representations by dynamically reconstructing diverse image contexts. This approach applies to both X-ray and CT modalities.

\textbf{Optimal Transport Theory.} OT determines an efficient way to transform one distribution into another, considering the associated costs or distances between individual elements in the distributions~\cite{chen2021wasserstein}. It is popular for computer vision tasks like domain adaptation~\cite{ding2023cross}, multitask learning~\cite{janati2019wasserstein}, feature matching~\cite{sarlin2020superglue}, generative model~\cite{bunne2019learning} etc. where measuring dissimilarity or similarity between distributions is essential. Zheng \textit{et al.} ~\cite{ge2021ota} framed label assignment in object detection as an OT problem where the unit transportation cost between anchor and ground-truth pairs was defined by weighted summation of classification and regression losses, solved through Sinkhorn-Knopp iteration for optimal assignment. Kim \textit{et al.} proposed an identity-invariant facial expression recognition method using OT to quantify inter-identity variation for optimal matching of similar expressions across different identities~\cite{kim2022optimal}. Zhu \textit{et al.} introduced an OT-guided translation network for unpaired image-to-image translation schemes to enhance retinal color fundus photography~\cite{zhu2023otre}. CUI \textit{et .al} in~\cite{yufei2023retrieval} leveraged OT for nearest neighbor retrieval in on whole slide images (WSIs) to address performance degradation in multiple instance learning models when tested on out-of-domain data. OT has limited applicability in the CXRs and medical domain due to complexities inherent in medical image data structures. In this work, we introduce a novel integration of OT within the SSL framework to enhance the effective handling of the CXR tasks.

\section{Proposed Method}
\label{sec:Method}
In this section, we first present preliminaries on OT theory and subsequently describe the proposed SSL framework OTCXR, which comprises two main stages: 1) formulation of \textit{dense semantic invariance} as an OT problem and 2) cross-viewpoint semantics infusion module (CV-SIM).

% \subsection{Preliminaries of OT}
% Optimal transportation compares two probability  distributions ($\mu$ and $\nu$) by determining the most efficient and cost-effective transportation plan $\pi^{*}$ to transform one distribution into the other. Let $\mu$ and $\nu$ be the be the probability distributions over the visual feature spaces $X$ and $Y$ such that:
% $\mu = \sum_{i}^{n}p_{s_i}\delta(x_i)$ and $\nu = \sum_{i}^{m}p_{t_j}\delta(y_j)$. $p_s{_i}$ and $p_{t_j}$ are probabilities associated with different visual features. The goal of OT is to find an optimal transport plan $\pi^{*}$ such that the total cost of transporting features from one image to another is minimum:
% \begin{equation}
% \begin{split}    
%     \pi^{*} = \argmin_{\pi \in \mathbb{R}_{+}^{n \times m}}\sum{i=1}^{n}\sum{j=1}^{m} \pi_{ij} C_{ij}
%     \\s.t. \sum_{i=1}^{n}\pi_{ij} = \mu(x_i)and \sum_{j=1}^{m}\pi_{ij} = \nu(y_j)
% \end{split}
% \end{equation} 
% Here, $\pi_{ij}$ represents the optimal amount of mass to move from feature $x_i$ in one image to feature $y_j$ in another image, while $C_{ij}$ is the cost matrix given by by a cost function $c(x_i,y_j)$ as the cost of transporting a feature from $x_i$ to $y_j$.

\textbf{OTCXR SSL Framework.}
OT has recently gained interest in computer vision applications that involve comparing images. By determining the optimal transportation plan between two images, we can quantify the similarity or dissimilarity between them in a way that considers the spatial arrangement of pixel intensities. This is particularly useful in CXR analysis, where the pixel-wise differences may not capture perceptual similarity. 

In this work, we adapt OT in SSL to measure the optimal distance between two sets of features representing the embeddings extracted from each image. The primary objective is to ascertain the optimal flow of information between source and target images while ensuring the minimum associated cost in terms of distance between features. 
More specifically, as illustrated in Figure~\ref{fig:workflow_1}, the OTCXR framework generates two augmented versions ($I_s$ and $I_t$) of the original input image $I$ by applying random augmentations $t_s$ and $t_t$ sampled from a predefined set of augmentations $\mathcal{T}$. \new{Following the literature~\cite{simclr,mococxr,byol}, we adopted data augmentations common in SSL pre-training. Particularly OTCXR employs random resized cropping, color jitter, random horizontal flipping, and Gaussian blur.}

$I_s$ and $I_t$ undergo processing through two weight-shared CNN encoders $f_s$ and $f_t$. The output of $f_s$ and $f_t$ are the dense feature maps $z_s \in \mathbb{R}^{d \times hw}$ and $z_t\in \mathbb{R}^{d \times hw}$ respectively, where $d$ represents the number of channels while $hw$ captures the spatial dimensions. Subsequently, $z_s$ and $z_t$ are utilized to compute the discrepancy matrix $C_{ij} \in \mathbb{R}^{d\times d}$, where each element quantifies the cosine similarity between feature vectors at position 
$i$ in $z_s$ and the feature vector at position $j$ in $z_t$. The discrepancy matrix $C_{ij}$ is constructed as follows:
\begin{equation}
    C_{ij} = \frac{z_s^T. z_t}{\left\|z_s\|\right.\left\|z_t\right\|}
\end{equation}
Dense feature maps $z_s$ and $z_t$ offer an advantage over post-pooling feature vectors for semantic alignment, as they preserve the spatial information and relationships between pixels in the original image. 
This enables the model to learn more meaningful representations for CXRs, where the spatial arrangement of structures is vital for accurate diagnosis. The formulation of the OT problem aligns seamlessly with dense feature maps $z_s$ and $z_t$, allowing for a more granular and accurate estimation of the cost associated with transferring information between corresponding positions. 

Furthermore, $z_s$ and $z_t$ undergo a global average pooling layer $GAP$ to output $g_s$ and $g_t$, which condenses the spatial information captured by the dense feature maps into single-dimensional compact representations. Subsequently, $g_s$ and $g_t$ are projected to a high dimensional space using $MLP$ expansion heads $h_s$ and $h_t$ to output $q_s$ and $q_t$ respectively. $g_s$ and $g_t$ 
are further utilized by the CV-SIM modules $R_s$ and $R_t$ in each branch of the network. The subsequent section provides the details of the CV-SIM. 

\textbf{Cross-Viewpoint Semantics Infusion Module.}
CV-SIM module incorporates a multi-head cross-view attention mechanism, where each head attends to diverse subspaces of queries and keys. This contributes to navigating the high-dimensional feature space to extract subtle relationships and dependencies across different viewpoints within the data. In this context, it linearly transform $g_s$ into $q$ and $g_t$ into $k$ and $v$, each having a dimension of $\mathbb{R}^{d}$. Subsequently, a multi-head attention mechanism computes attention scores $\hat{a}$ through a scaled dot-product between queries ($q$) and keys ($k$) by employing $\sigma(\frac{q.k^{T}}{\sqrt{dim}})$. $\sigma$ represents the softmax function, which transforms relevant attention scores into attention weights, while $dim$ denotes the dimensionality of the query and key vector.

Subsequently, CV-SIM refines the representation by computing the attended values through a weighted sum operation, where the attention weights guide the aggregation of information from the target representation $v$ based on the computed attention weights $\hat{a}$. This ensures that CV-SIM captures intricate dependencies across different viewpoints, contributing to the enrichment of the overall feature space. The attended values computation yields an enhanced representation $r_s \in \mathbb{R}^d$. Likewise, a corresponding procedure is executed in $R_t$, where the roles of $g_s$ and $g_t$ are interchanged to generate the corresponding outcome $r_t \in \mathbb{R}^d$. The resulting enhanced representations $r_s$ and $r_t$ from $R_s$ and $R_t$, respectively, \new{undergo} another transformation through the softmax function to convert it into a probabilistic distribution \new{$\mu$ and $\nu$} over the visual feature spaces respectively.
 
OT solver utilized this \new{$\mu$ and $\nu$} as the marginal constraints of the transport plan such that the total cost of transporting features from one image to another is minimal.  This step ensures that the probability distribution derived from the refined representation is effectively incorporated into the optimization process facilitated by the OT solver, contributing to the alignment of features across different branches within the OTCXR framework.

\begin{table*}[t!]
\centering
\caption{\new{Performance comparison of SSL methods and OTCXR for chest X-ray classification on NIH, VinBig-CXR, and RSNA datasets. Results show AUC scores (\%) for NIH and VinBig-CXR and accuracy (\%) for RSNA under semi-supervised linear evaluation, using 1\%, 10\%, and 100\% of labeled training data. Values in parentheses indicate performance differences from the supervised baseline (Sup.2). Arrows $\uparrow$ and $\downarrow$ denote improvement or decrease, respectively; `-' indicates no change.}}

\label{tab:xray}
\adjustbox{width=1.0\textwidth}{          
\begin{tabular}{ccccccccccccc}
\toprule
\multirow{3}{*}{\textbf{Methods}} & \multicolumn{3}{c}{\textbf{NIH}} & \multicolumn{3}{c}{\textbf{VinBig-CXR}} & \multicolumn{3}{c}{\textbf{RSNA}}  \\ \cmidrule(lr){2-4} \cmidrule(lr){5-7} \cmidrule(lr){8-10} 
& 1\% & 10\% & 100\%  & 1\% & 10\% & 100\% &1\% &10\% & 100\%  \\   
\midrule
\midrule  
\rowcolor{lightgray!15}\multicolumn{1}{c}{Sup.1}  
&$61.0_{(-6.8)}\textcolor{myred}\downarrow $
&$71.1_{(-3.5)}\textcolor{myred}\downarrow $
&$76.8_{(-4.1)}\textcolor{myred}\downarrow $
&$71.1_{(-8.7)}\textcolor{myred}\downarrow $
&$85.0_{(-3.3)}\textcolor{myred}\downarrow $
&$91.0_{(-2.4)}\textcolor{myred}\downarrow $
&$77.0_{(-1.9)}\textcolor{myred}\downarrow $
&$79.8_{(-1.1)}\textcolor{myred}\downarrow $
&$80.0_{(-3.1)}\textcolor{myred}\downarrow $ \\
\rowcolor{gray!20}\multicolumn{1}{c}{$Sup.2_{(bs)}$} &$67.8_{(0.0)}$
&$74.6_{(0.0)}$
&$80.9_{(0.0)}$
&$79.8_{(0.0)}$
&$88.3_{(0.0)}$
&$93.4_{(0.0)}$
&$78.9_{(0.0)}$
&$80.9_{(0.0)}$
&$83.1_{(0.0)}$\\
\midrule
\midrule
\rowcolor{mydeepblue!3}\multicolumn{1}{c}{SimCLR} 
&$67.1_{(-0.7)}\textcolor{myred}\downarrow $
&$74.8_{(0.2)}\textcolor{mygreen}\uparrow $
&$78.9_{(-2.0)}\textcolor{myred}\downarrow $
&$77.9_{(-1.9)}\textcolor{myred}\downarrow $
&$87.7_{(-0.6)}\textcolor{myred}\downarrow $
&$92.4_{(-1.0)}\textcolor{myred}\downarrow $
&$80.0_{(1.1)}\textcolor{mygreen}\uparrow $
&$81.1_{(0.2)}\textcolor{mygreen}\uparrow $
&$81.8_{(-1.3)}\textcolor{myred}\downarrow $\\

\rowcolor{mydeepblue!8}\multicolumn{1}{c}{MoCoV2} 
&$66.8_{(-1.0)}\textcolor{myred}\downarrow $
&$74.6_{(0.0)}\textcolor{black}-$
&$79.4_{(-1.5)}\textcolor{myred}\downarrow $
&$77.1_{(-2.7)}\textcolor{myred}\downarrow $
&$86.2_{(-2.1)}\textcolor{myred}\downarrow $
&$92.0_{(-1.4)}\textcolor{myred}\downarrow $
&$79.4_{(0.5)}\textcolor{mygreen}\uparrow $
&$81.3_{(0.4)}\textcolor{mygreen}\uparrow $
&$82.0_{(-1.1)}\textcolor{myred}\downarrow $\\

\rowcolor{mydeepblue!13}\multicolumn{1}{c}{DenseCL} 
& $67.7_{(-0.1)}\textcolor{myred}\downarrow $
& $76.4_{(1.8)}\textcolor{mygreen}\uparrow $
& $81.4_{(0.5)}\textcolor{mygreen}\uparrow $
& $82.7_{(2.9)}\textcolor{mygreen}\uparrow $
& $88.7_{(0.4)}\textcolor{mygreen}\uparrow $
& $93.9_{(0.5)}\textcolor{mygreen}\uparrow $
& $79.9_{(1.0)}\textcolor{mygreen}\uparrow $
& $81.1_{(0.2)}\textcolor{mygreen}\uparrow $
& $82.3_{(-0.8)}\textcolor{myred}\downarrow $ \\ 
\midrule
\midrule
\rowcolor{mydeepblue!3}\multicolumn{1}{c}{BYOL} 
&$66.3_{(-1.5)}\textcolor{myred}\downarrow $
&$74.5_{(-0.1)}\textcolor{myred}\downarrow $
&$78.8_{(-2.1)}\textcolor{myred}\downarrow $
&$76.2_{(-3.6)}\textcolor{myred}\downarrow $
&$85.8_{(-2.5)}\textcolor{myred}\downarrow $
&$85.9_{(-7.5)}\textcolor{myred}\downarrow $
&$78.3_{(-0.6)}\textcolor{myred}\downarrow $
&$80.5_{(-0.4)}\textcolor{myred}\downarrow $
&$82.5_{(-0.6)}\textcolor{myred}\downarrow $ \\
\rowcolor{mydeepblue!8}\multicolumn{1}{c}{SimSiam} 
&$66.6_{(-1.2)}\textcolor{myred}\downarrow $
&$74.3_{(-0.3)}\textcolor{myred}\downarrow $
&$78.5_{(-2.4)}\textcolor{myred}\downarrow $
&$75.8_{(-4.0)}\textcolor{myred}\downarrow $
&$85.9_{(-2.4)}\textcolor{myred}\downarrow $
&$91.5_{(-1.9)}\textcolor{myred}\downarrow $
&$78.1_{(-0.8)}\textcolor{myred}\downarrow $
&$80.1_{(-0.8)}\textcolor{myred}\downarrow $
&$82.1_{(-1.0)}\textcolor{myred}\downarrow $ \\
\rowcolor{mydeepblue!13}\multicolumn{1}{c}{VICReg} 
&$67.4_{(-0.4)}\textcolor{myred}\downarrow $
&$75.5_{(0.9)}\textcolor{mygreen}\uparrow $
&$80.2_{(-0.7)}\textcolor{myred}\downarrow $
&$76.9_{(-2.9)}\textcolor{myred}\downarrow $
&$87.5_{(-0.8)}\textcolor{myred}\downarrow $
&$91.9_{(-1.5)}\textcolor{myred}\downarrow $
&$79.5_{(0.6)}\textcolor{mygreen}\uparrow $
&$80.7_{(-0.2)}\textcolor{myred}\downarrow $
&$82.7_{(-0.4)}\textcolor{myred}\downarrow $ \\
\rowcolor{mydeepblue!18}\multicolumn{1}{c}{PCRLv2} 
& $67.5_{(-0.3)}\textcolor{myred}\downarrow $
& $75.6_{(1.0)}\textcolor{mygreen}\uparrow $
& $81.0_{(0.1)}\textcolor{mygreen}\uparrow $
& $75.9_{(-3.9)}\textcolor{myred}\downarrow $
& $87.8_{(-0.5)}\textcolor{myred}\downarrow $
& $93.4_{(0.0)}\textcolor{black}-$
& $80.1_{(1.2)}\textcolor{mygreen}\uparrow $
& $81.5_{(0.6)}\textcolor{mygreen}\uparrow $
& $82.9_{(-0.2)}\textcolor{myred}\downarrow $ \\
\midrule
\midrule
\rowcolor{mydeepblue!18}\multicolumn{1}{c}{\textbf{OTCXR}}&
$ \textbf{68.8}_{(1.0)}\textcolor{mygreen}\uparrow $
& $\textbf{77.6}_{(3.0)}\textcolor{mygreen}\uparrow $
& $\textbf{82.5}_{(1.6)}\textcolor{mygreen}\uparrow $
& $\textbf{83.8}_{(4.0)}\textcolor{mygreen}\uparrow $
& $\textbf{89.5}_{(1.2)}\textcolor{mygreen}\uparrow $
& $\textbf{94.6}_{(1.2)}\textcolor{mygreen}\uparrow $
& $\textbf{81.4}_{(2.5)}\textcolor{mygreen}\uparrow $
& $\textbf{82.3}_{(1.4)}\textcolor{mygreen}\uparrow $
& $\textbf{84.0}_{(0.9)}\textcolor{mygreen}\uparrow $ \\

\bottomrule
\end{tabular} }
\end{table*}

\textbf{Modelling Dense Semantic Invariance as OT Problem.}
In the OTCXR framework, we cast the challenge of achieving dense semantic invariance between $I_s$ and $I_t$ as an OT problem by employing a matrix $T$ representing the matching distribution of features across different viewpoints. The goal is to maximize semantic invariance across $I_s$ and $I_t$ by obtaining a global optimal transport plan $T^{*}$. To formulate the optimization problem, we define the total discrepancy as $\sum_{ij}T_{ij}C_{ij}$ where $C_{ij}$ is the cosine similarity-based discrepancy matrix computed from the dense feature $z_s$ and $z_t$. To find $T^{*}$ that maximizes this total discrepancy to ensure effective alignment of semantic features, we define the cost matrix $M_{ij} = 1 - C_{ij}$, which we need to minimize. To avoid the trivial solutions, we introduce probability distributions $\mu$ and $\nu$ in the CV-SIM of the OTCXR framework. Note that in the CV-SIM module, $\mu$ and $\nu$ are not simply initialized as uniform distributions, which is a common practice, but instead, these are computed using the intra- and inter-relational information. This means that $\mu$ and $\nu$ encode task-specific information while giving less weight to the irrelevant dimensions. The OT problem is then formulated as follows:
% , which is a less expensive way to learn $T^{*}$.
\begin{equation}
     T^{*} = \argmin_{T \in \prod(\mu,\nu)}
    \sum_{i=1}^{d} \sum_{j=1}^{d} T_{ij} M_{ij} = \min_{T}\langle T,M \rangle
    \label{eq:wd}
\end{equation}
where $\prod(\mu,\nu)$ represents the set of transport plans subjected to $T \in \mathbb{R}^{d\times d}_+: T \mathbf{1} = \mu$,  $T^\top \mathbf{1} = \nu$. The constraints $T \mathbf{1} = \mu$ and $T^\top \mathbf{1} = \nu$ enforce the conservation of mass where $\mathbf{1}$ denotes a vector of ones. $T_{ij}$ is the amount of mass moved from $i^{th}$ point in $\mu$ to $j^{th}$ point in $\nu$ to achieve minimal cost. The optimization process formulated in equation~\eqref{eq:wd} seeks to determine the optimal transport plan $T^{*}$ that minimizes the total matching difference subjected to prescribed marginal constraints. Equation~\eqref{eq:wd} can be efficiently solved using Sinkhorn algorithm~\cite{cuturi2013sinkhorn,peyre2019computational}, which makes it compatible with the deep learning framework. $T^{*}$ is subsequently used to obtain the OT loss term for the final objective as $L_{OT} = \langle T^*, M\rangle$.

Further, inspired by VICReg~\cite{vicreg}, we employ two regularization terms: Variance ($var$) and Covariance ($cov$), using $q_s$ and $q_t$ independently in both branches of the model's architecture mainly to avoid any possibility of collapse. The final objective for OTCXR is formulated in Eq.~\eqref{eq:final}, where $\alpha$, $\beta$, and $\eta$, are hyper-parameters. 
\begin{equation}
\begin{split}
L_{MT} = \alpha \times L_{OT} + \beta \times[var(q_s) + var(q_t)] \\
    + \eta \times [cov(q_s) + cov(q_t)]
\end{split}
\label{eq:final}
\end{equation}

By iteratively refining the representations through CV-SIM and aligning them using OT, the method establishes a mechanism that inherently promotes dense semantic invariance while fostering view-invariant features. The result is a set of representations that capture not only the details specific to each view but also the essential semantic content that is consistent across views.

\section{Experimental Platform} \label{sec:results}
\textbf{Datasets.}
The proposed approach employs SSL pre-training using publicly available NIH Chest X-ray~\cite{wang2017chestx} dataset, which has 112,120 chest X-ray images labeled with 14 thoracic pathologies. Further, for evaluations of the learned representations, we use VinBig-CXR~\cite{nguyen2022vindr} and RSNA Pneumonia~\cite{rsna} datasets along with the NIH dataset.
VinBig-CXR contains 18K chest X-ray images, annotated into 14 pathologies. The RSNA~\cite{rsna} dataset consists of 30k chest radiographs captured from a frontal view, each annotated as either healthy or pneumonia. We also perform segmentation as the downstream evaluation on SIIM-ACR Pneumothorax~\cite{siim} data.

\textbf{Evaluation Protocols} 
To assess the efficacy of the learned representations, we conduct downstream tasks involving chest X-ray image classification and segmentation using only the backbone encoder $f_s$ obtained from pre-training. Following the established practices in the literature\cite{simclr,byol,pcrlv2,vicreg},  we adopted two distinct evaluation protocols (1) frozen and (2) fine-tuning for which we add a single linear layer classifier on top of the pre-trained backbone encoder. In the frozen evaluation protocol, the parameters of the backbone CNN encoder remain fixed, ensuring that only the linear layer's parameters are updated during downstream training. Conversely, in the fine-tuning evaluation protocol, we fine-tuned the SSL pre-trained backbone CNN encoder and the newly added linear classifier layer. We presented results on the test/validation dataset using various subsets (1\%, 10\%, and All) of the training data. We perform the segmentation under fine-tuning evaluation protocol using Resent-based U-Net architecture after updating the encoder's parameter with that obtained from OTCXR's pre-training. 

\textbf{Method Compared.} For comparison, the baselines encompassed both supervised, including initialization with random (Sup.1) and ImageNet (Sup.2) weights and a range of SOTA SSL techniques.  In the SSL domain, we included as SimCLR~\cite{simclr}, MoCov2~\cite{mocov2}, BYOL~\cite{byol}, SimSiam~\cite{simsiam}, VICReg~\cite{vicreg} and PCRLv2~\cite{pcrlv2}. We perform the pre-training for these baselines ourselves, following their official implementations and aligning with our proposed approach's training protocol.

\textbf{Implementation Details}
We utilize ResNet18 as the backbone encoder for SSL pre-training. The training of the encoder involves a batch size of $64$ over a span of $300$ epochs. The encoder output is then propagated through an MLP head serving as an expander. This expander head comprises three linear layers, each featuring a dimensionality of 2048, and is accompanied by ReLU activation and batch normalization within each layer. The optimization strategy employed is the LARS optimization algorithm, with a learning rate of $3e-4$ and a weight decay of $1e-4$. The values of $\beta$ and $\eta$ are directly adopted from VICReg~\cite{vicreg}, while $\alpha$ is empirically set to $0.6$ for optimal performance.

\begin{table*}[t!]
\centering
% \caption{Linear Evaluation on different datasets and subsets under frozen settings. The table compares the proposed method and baselines across NIH Chest X-ray, Vinbig-CXR, and RSNA datasets.
% }

\caption{\new{Performance comparison of SSL methods for chest X-ray classification under linear evaluation with frozen features. Results show AUC scores (\%) for NIH and VinBig-CXR datasets and accuracy (\%) for the RSNA dataset, using 1\%, 10\%, and 100\% of labeled training data. Values in parentheses indicate performance differences from the supervised baseline (Sup.2). Arrows $\uparrow$ and $\downarrow$ denote improvement or decrease, respectively; `-' indicates no change. This evaluation demonstrates the quality of learned representations without fine-tuning.}}

\label{tab:xray2}
\adjustbox{width=1.0\textwidth}{           
\begin{tabular}{cccccccccccccccc}
\toprule
\multirow{3}{*}{\textbf{Methods}} & \multicolumn{3}{c}{\textbf{NIH}} & \multicolumn{3}{c}{\textbf{Vinbig-CXR}} & \multicolumn{3}{c}{\textbf{RSNA}} \\ \cmidrule(lr){2-4} \cmidrule(lr){5-7} \cmidrule(lr){8-10}

& 1\% & 10\% &100\% & 1\%  & 10\% & 100\% & 1\% & 10\% &100\%
\\   
\midrule
\rowcolor{lightgray!15}\multicolumn{1}{c}{Sup.1}  
&$56.6_{(-6.0)}\textcolor{myred}\downarrow $
&$59.1_{(-9.2)}\textcolor{myred}\downarrow $
&$61.8_{(-11.0)}\textcolor{myred}\downarrow $
&$56.8_{(-0.6)}\textcolor{myred}\downarrow $
&$59.4_{(-16.8)}\textcolor{myred}\downarrow $
&$71.1_{(-14.0)}\textcolor{myred}\downarrow $
&$75.7_{(-1.7)}\textcolor{myred}\downarrow $
&$76.8_{(-2.6)}\textcolor{myred}\downarrow $
&$78.2_{(-2.2)}\textcolor{myred}\downarrow $ \\

\rowcolor{gray!15}\multicolumn{1}{c}{$Sup.2_{(bs)}$}  
&$62.6_{(0.0)}$
&$68.3_{(0.0)} $
&$72.8_{(0.0)}$
&$57.4_{(0.0)}$
&$76.2_{(0.0)}$
&$85.1_{(0.0)}$ 
&$77.4_{(0.0)} $
&$79.4_{(0.0)} $
&$80.4_{(0.0)}$  \\
\midrule

\rowcolor{mydeepblue!5}\multicolumn{1}{c}{SimCLR}  
&$62.4_{(-0.2)}\textcolor{myred}\downarrow $
&$68.3_{(0.0)}-$
&$73.0_{(0.2)}\textcolor{mygreen}\uparrow $
&$66.3_{(8.9)}\textcolor{mygreen}\uparrow $
&$77.4_{(1.2)}\textcolor{mygreen}\uparrow $
&$85.3_{(0.2)}\textcolor{mygreen}\uparrow $
&$76.9_{(-0.5)}\textcolor{myred}\downarrow $
&$78.9_{(-0.5)}\textcolor{myred}\downarrow $
&$81.8_{(1.4)}\textcolor{mygreen}\uparrow $ \\

\rowcolor{mydeepblue!8}\multicolumn{1}{c}{MoCov2} 
&$61.9_{(-0.7)}\textcolor{myred}\downarrow $
&$68.9_{(0.6)}\textcolor{mygreen}\uparrow $
&$72.6_{(-0.2)}\textcolor{myred}\downarrow $
&$66.7_{(9.3)}\textcolor{mygreen}\uparrow $
&$74.6_{(-1.6)}\textcolor{myred}\downarrow $
&$85.5_{(0.4)}\textcolor{mygreen}\uparrow $
&$76.8_{(-0.6)}\textcolor{myred}\downarrow $
&$78.1_{(-1.3)}\textcolor{myred}\downarrow $
&$81.2_{(0.8)}\textcolor{mygreen}\uparrow $ \\

\rowcolor{mydeepblue!13}\multicolumn{1}{c}{DenseCL} 
&$62.9_{(0.3)}\textcolor{mygreen}\uparrow $
&$69.5_{(1.2)}\textcolor{mygreen}\uparrow $
&$72.7_{(-0.1)}\textcolor{myred}\downarrow $
&$64.3_{(6.9)}\textcolor{mygreen}\uparrow $
&$73.6_{(-2.6)}\textcolor{myred}\downarrow $
&$85.5_{(0.4)}\textcolor{mygreen}\uparrow $
&$76.8_{(-0.6)}\textcolor{myred}\downarrow $
&$78.8_{(-0.6)}\textcolor{myred}\downarrow $
&$81.2_{(0.8)}\textcolor{mygreen}\uparrow $ \\
\midrule

\rowcolor{mydeepblue!5}\multicolumn{1}{c}{BYOL}  
&$61.7_{(-0.9)}\textcolor{myred}\downarrow $
&$65.8_{(-2.5)}\textcolor{myred}\downarrow $
&$69.9_{(-2.9)}\textcolor{myred}\downarrow $
&$65.6_{(8.2)}\textcolor{mygreen}\uparrow $
&$77.8_{(1.6)}\textcolor{mygreen}\uparrow $
&$84.0_{(-1.1)}\textcolor{myred}\downarrow $
&$76.5_{(-0.9)}\textcolor{myred}\downarrow $
&$78.4_{(-1.0)}\textcolor{myred}\downarrow $
&$80.9_{(0.5)}\textcolor{mygreen}\uparrow $ \\

\rowcolor{mydeepblue!8}\multicolumn{1}{c}{SimSiam} 
&$62.0_{(-0.6)}\textcolor{myred}\downarrow $
&$65.9_{(-2.4)}\textcolor{myred}\downarrow $
&$69.8_{(-3.0)}\textcolor{myred}\downarrow $
&$65.0_{(7.6)}\textcolor{mygreen}\uparrow $
&$77.5_{(1.3)}\textcolor{mygreen}\uparrow $
&$84.1_{(-1.0)}\textcolor{myred}\downarrow $
&$76.2_{(-1.2)}\textcolor{myred}\downarrow $
&$77.8_{(-1.6)}\textcolor{myred}\downarrow $
&$80.5_{(0.1)}\textcolor{mygreen}\uparrow $ \\

\rowcolor{mydeepblue!13}\multicolumn{1}{c}{VicReg} 
& $62.7_{(0.1)}\textcolor{mygreen}\uparrow $
& $68.7_{(0.4)}\textcolor{mygreen}\uparrow $
& $72.3_{(-0.5)}\textcolor{myred}\downarrow $
& $65.3_{(7.9)}\textcolor{mygreen}\uparrow $
& $76.5_{(0.3)}\textcolor{mygreen}\uparrow $
& $83.2_{(-1.9)}\textcolor{myred}\downarrow $
& $76.7_{(-0.7)}\textcolor{myred}\downarrow $
& $78.4_{(-1.0)}\textcolor{myred}\downarrow $
& $81.2_{(0.8)}\textcolor{mygreen}\uparrow $ \\

\rowcolor{mydeepblue!3}\multicolumn{1}{c}{PCRLv2}
&$59.8_{(-2.8)}\textcolor{myred}\downarrow $
&$68.6_{(0.3)}\textcolor{mygreen}\uparrow $
&$72.9_{(0.1)}\textcolor{mygreen}\uparrow $
&$59.8_{(2.4)}\textcolor{mygreen}\uparrow $
&$71.9_{(-4.3)}\textcolor{myred}\downarrow $
&$78.0_{(-7.1)}\textcolor{myred}\downarrow $
&$76.8_{(-0.6)}\textcolor{myred}\downarrow $
&$80.3_{(0.9)}\textcolor{mygreen}\uparrow $
&$81.1_{(0.7)}\textcolor{mygreen}\uparrow $ \\

\midrule
\rowcolor{mydeepblue!18}\multicolumn{1}{c}{\textbf{OTCXR}} 
& $\textbf{63.8}_{(1.2)}\textcolor{mygreen}\uparrow $
& $\textbf{70.3}_{(2.0)}\textcolor{mygreen}\uparrow $
& $\textbf{73.9}_{(1.1)}\textcolor{mygreen}\uparrow $
& $\textbf{66.1}_{(8.7)}\textcolor{mygreen}\uparrow $
& $\textbf{78.1}_{(1.9)}\textcolor{mygreen}\uparrow $
& $\textbf{86.5}_{(1.4)}\textcolor{mygreen}\uparrow $
& $\textbf{80.0}_{(2.6)}\textcolor{mygreen}\uparrow $
& $\textbf{81.3}_{(1.9)}\textcolor{mygreen}\uparrow $
& $\textbf{82.1}_{(1.7)}\textcolor{mygreen}\uparrow $ \\

\bottomrule
\end{tabular}}
\end{table*}

\begin{figure*}[t!]
\centering
\begin{tabular}{ccccccccccc}
OTCXR \hspace{1.0cm} PCRLv2 \hspace{1.0cm} VICReg \hspace{1.0cm}  DenseCL \hspace{1.1cm}  BYOL \hspace{1.1cm} SimSiam \hspace{1.0cm}  SimCLR \\
\multicolumn{11}{l}{\includegraphics[width=0.95\linewidth]{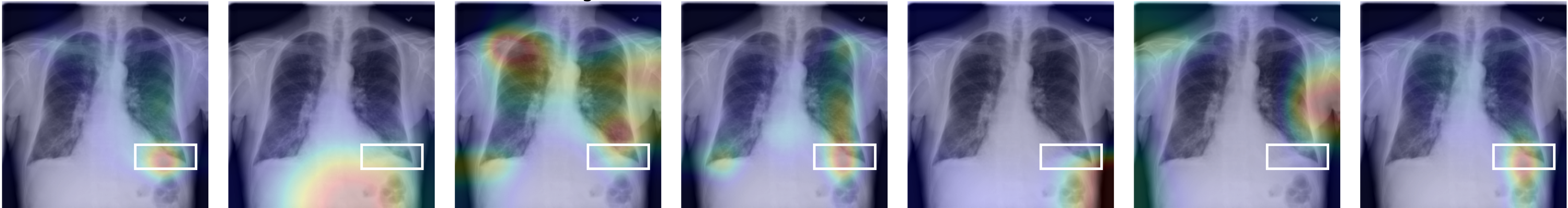}} \\
\includegraphics[width=0.95\linewidth]{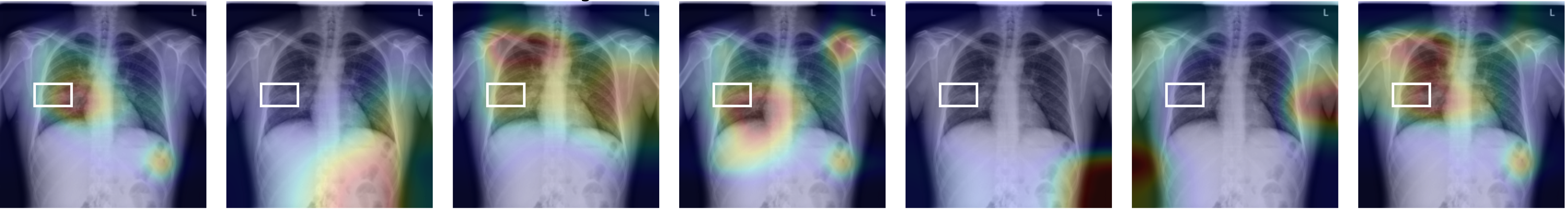} \\
\includegraphics[width=0.95\linewidth]{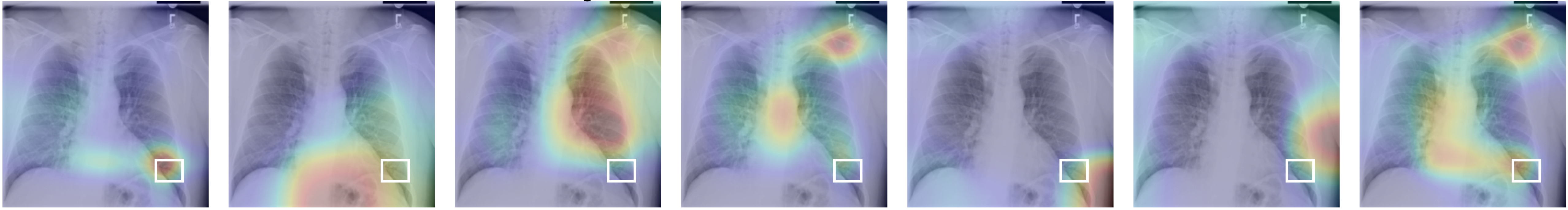} \\
\includegraphics[width=0.95\linewidth]{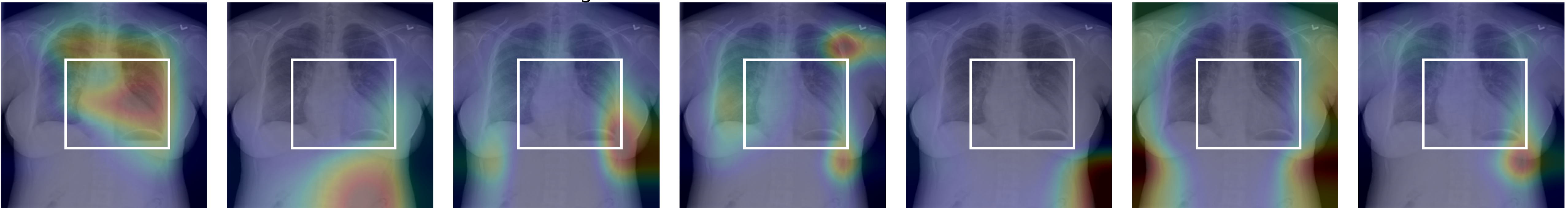} \\
\end{tabular}
\caption[short]{Diagnostic heat maps generated by OTCXR and the considered SSL baseline methods represent interpretations of chest X-ray images fine-tuned with 1\% of training samples from the NIH dataset.}
\label{fig_cam:gradcam}
\end{figure*} 

\subsection{Results and Analysis}
This section presents the evaluation results of the proposed framework under the following evaluation protocols.

\subsubsection{Quantitative Results under Finetuning Protocol} 
Table~\ref{tab:xray} presents the evaluation results from the test set on the downstream classification task under liner probing with fine-tuning using subsets (1\%, 10\%, and 100\%) of labeled data from the train set. 
OTCXR consistently outperforms the baselines across all the evaluated tasks \new{and datasets}. \new{For the NIH dataset, OTCXR achieves the highest AUC scores of 68.8\%, with notable improvements of 1.0\%, 3.0\%, and 1.6\% over the supervised baseline for 1\%, 10\%, and 100\% labeled data, respectively. Compared to SSL baselines on NIH, OTCXR shows consistent improvements. For instance, with 1\% labeled data, OTCXR outperforms the next best SSL method (DenseCL) by 1.1\%. The gap widens with 10\% labeled data, 
where OTCXR consistently surpasses the considered SSL baselines. With 100\% labeled data, OTCXR maintains its lead, outperforming the best-performing baseline method (PCRLv2) by 1.5\%. Notably, OTCXR outperforms all SOTA SSL methods, including PCRLv2, specifically designed for CXR analysis and supervised baselines across different labeled data proportions. This consistent superiority, especially in scenarios with limited labeled data (1\% and 10\%), underscores OTCXR's effectiveness in learning robust and transferable representations for chest X-ray analysis. }

\textbf{Transferability to Other Datasets: }
In transfer learning, OTCXR consistently exhibits performance gain \new{on both the VinBig-CXR and RSNA datasets. For VinBig-CXR, OTCXR achieves AUC score improvements of 4.0\%, 1.2\%, and 1.2\% over the supervised baseline for 1\%, 10\%, and 100\% labeled data scenarios, respectively. Similarly, on the RSNA dataset, OTCXR outperforms all SSL and supervised baselines, achieving the highest accuracy scores of 81.4\%, 82.3\%, and 84.0\% for 1\%, 10\%, and 100\% labeled data, representing improvements of 2.5\%, 1.4\%, and 0.9\% over the baseline.} The performance gain indicates OTCXR's ability to learn generalized representations, demonstrating its potential for transfer learning scenarios where labeled samples are limited. The transferability of learned representations is particularly crucial in medical imaging applications where adapting models to new datasets or tasks is common, and OTCXR's performance gains on different datasets highlight its effectiveness in capturing versatile and transferable representations.

\subsubsection{Quantitative Results under Linear Evaluation(Frozen)}
Table~\ref{tab:xray2} reports the linear evaluation results on various datasets and subsets under frozen settings. In the case of the NIH Chest X-ray dataset, OTCXR consistently outperforms the baselines across different subsets, achieving an AUC score of \new{63.8\%} for 1\% labeled data. Similar trends are observed when labeled data increases to 10\% and when considered the whole data for which OTCXR obtained the highest AUC score of \new{73.9\%} by outperforming all the baseline methods. Similarly, on the Vinbig-CXR dataset, OTCXR excels across all subsets, attaining the highest AUC of \new{78.1\%} for the 10\% subset. However, for the 1\% subset, MoCoV2 exceptionally outperforms the proposed approach. For the RSNA dataset also, OTCXR \new{continues performs} better than the considered baseline methods. This reinforces its efficiency as a feature extractor in transfer learning scenarios, where efficient use of resources is a crucial consideration. 

\subsubsection{Qualitative Results}
Figure~\ref{fig_cam:gradcam} \new{presents diagnostic heat maps generated by OTCXR and SSL baseline methods, serving as visual representations of model interpretations for chest X-ray images. These models are fine-tuned using 1\% of training samples from the NIH dataset.}
The heat maps highlight regions within the chest X-ray images that are considered significant by the respective models for diagnostic purposes. \new{Upon comparison, we observe that OTCXR outperforms baseline SSL methods in accurately identifying regions of interest, as indicated by the ground truth bounding boxes. The heat maps generated by OTCXR demonstrate relatively more precise alignment with these diagnostically significant areas than those produced by baseline methods. Such performance showcases OTCXR's ability to leverage SSL techniques for enhanced interpretability in chest X-ray analysis, potentially contributing to more accurate and reliable clinical assessments.}

\begin{figure}[htbp]
    \centering
    \includegraphics[width=1.0\columnwidth,scale=0.3]{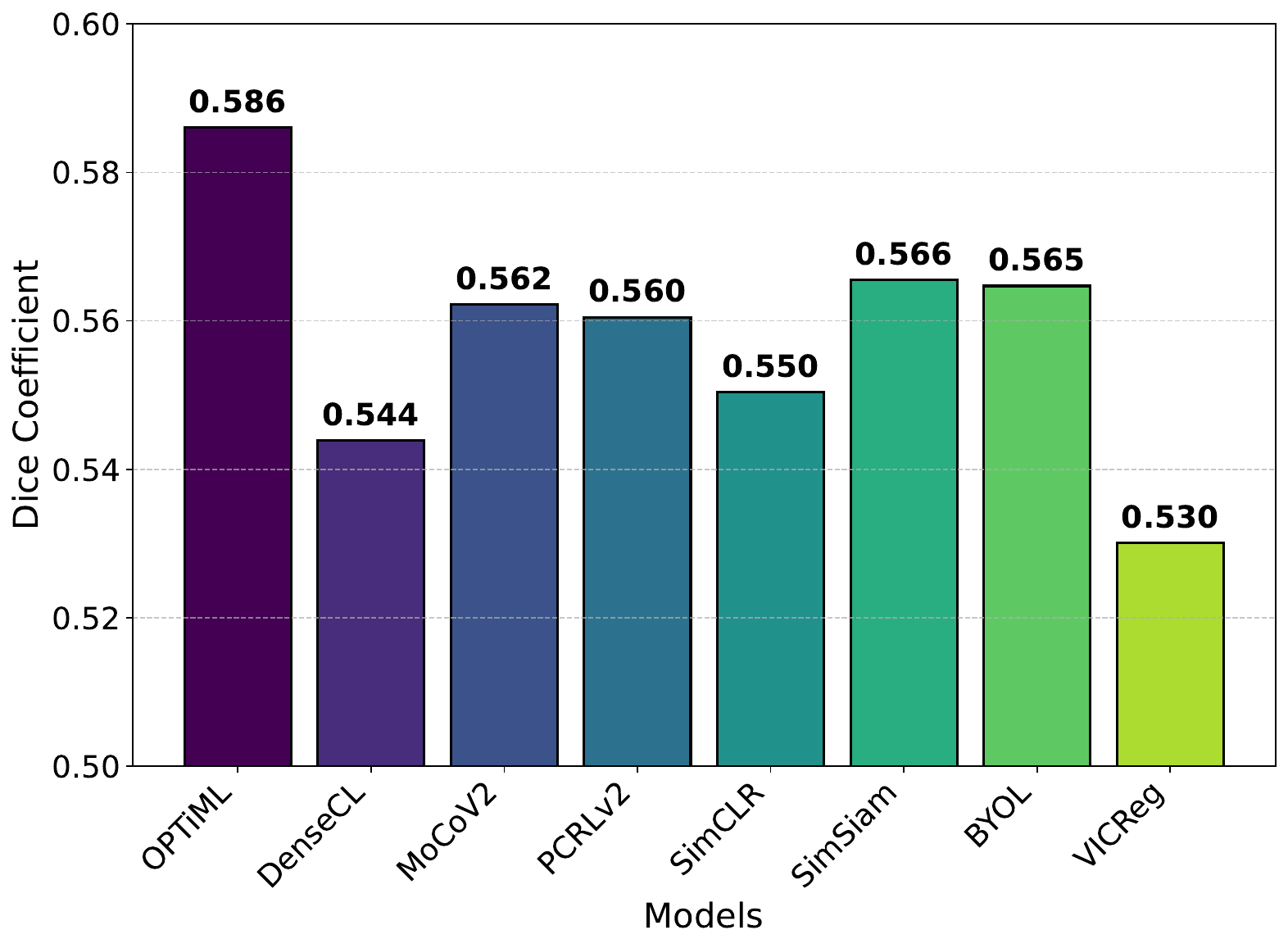}
    \caption{Segmentation on SIIM-ACR Pneumothorax dataset obtained after fine-tuning the representations obtained from NIH pre-training.}
    \label{fig:siim_acr_pneumothorax}
\end{figure}
\textbf{Transferability to Segmentation Task}
Furthermore, we also measure the performance of OTCXR under the transfer learning regime for segmentation as the downstream task.
For this, we consider the pneumothorax segmentation task on SIIM-ACR samples. Figure~\ref{fig:siim_acr_pneumothorax} presents the results for which we fine-tune a U-Net model which uses the backbone encoder initialized with the weights obtained from the OTCXR pre-training while the decoder is randomly initialized. The results in terms of the dice coefficient demonstrate a comparatively better generalization ability and transferability of the representations produced by the OTCXR, with the highest dice score of 0.586. Results show that dense semantic invariance, employed in OTCXR, is more effective than the traditional invariance approaches, such as in the baseline SSL methods.\\

\subsection{Ablation Study}
We conducted two studies to evaluate the effectiveness of different components of the OTCXR framework across various data percentages on the NIH dataset, and the observations are summarized in Table~\ref{tab:ablation}.

\textbf{Effect of $var$ and $cov$ Regularizers.} In this study, OTCXR is evaluated without the $var$ and $cov$ regularization components. The results indicate a decline in performance, highlighting the importance of these regularization terms in preserving valuable information during the training process. 

\textbf{Relevance of CV-SIM Module.} This study involves assessing OTCXR without the CV-SIM. This module is an important part of OTCXR, enhancing the model's ability to capture subtle dependencies across different viewpoints. The observed decrease in performance highlights its significance in maintaining and enriching the overall feature space.  Further, we discarded both $var$ and $cov$ regularization and the CV-SIM module from OTCXR. The combined absence of these components leads to a further decline in performance, emphasizing the cumulative impact of these elements on the overall effectiveness of the OTCXR framework. The results demonstrate that the optimal configuration, where both $var$ and $cov$ regularization and the CV-SIM module are included in OTCXR, yields the highest performance

\begin{table}[t!]
\centering
\caption{Results of an ablation study conducted on OTCXR across various data percentages on the NIH dataset. Values are presented as mean $\pm$ standard deviation over three runs.}
\adjustbox{width=1.0\columnwidth}{ 
\begin{tabular}{cccccc}
\toprule
Method &var+cov &CV-SIM &1\% & 10\% & 100\%\\
\midrule
\multirow{4}{*}{OTCXR}  & \xmark & \cmark  & 67.9 $\pm$ 0.3 & 76.5 $\pm$ 0.2 & 81.6 $\pm$ 0.1 \\
& \cmark  & \xmark & 67.8 $\pm$ 0.2 & 76.9 $\pm$ 0.3 & 81.4 $\pm$ 0.2 \\
& \xmark  & \xmark & 67.3 $\pm$ 0.3 & 76.2 $\pm$ 0.2 & 80.2 $\pm$ 0.1 \\
\rowcolor{mydeepblue!18}& \cmark  & \cmark & \textbf{68.8 $\pm$ 0.2} & \textbf{77.6 $\pm$ 0.1} & \textbf{82.5 $\pm$ 0.1} \\
\bottomrule
\end{tabular}
}
\label{tab:ablation}
\end{table}

\new{\subsection{Limitations and Computational Overhead}
While OTCXR demonstrates significant improvements in chest X-ray analysis, its potential limitation lies in the sensitivity of the OT solution to the marginal constraints derived from CV-SIM. This sensitivity may impact performance in rare or subtle chest X-ray pathologies where the global semantic context might not fully capture critical local features. Furthermore, the framework's effectiveness may vary across different imaging protocols or equipment, potentially necessitating fine-tuning for optimal performance in diverse clinical settings. Regarding computational aspects, OTCXR introduces minimal overhead due to the efficient Sinkhorn algorithm for solving the OT problem. This additional computation is confined to the pre-training phase, with no extra costs incurred during downstream tasks or inference. From a memory perspective, OTCXR adds $R_s$ and $R_t$ heads compared to existing SSL methods, resulting in a modest increase in memory requirements. These trade-offs in computational resources are justified by the significant performance improvements observed across various chest X-ray analysis tasks.}

\section{Conclusion}
In this work, we introduce OTCXR to enhance the capabilities of the SSL framework. The proposed framework demonstrates the effective integration of the OT in SSL to learn dense-semantic invariant features critical in CXRs. We also proposed a CV-SIM module to refine the dense semantic features from different viewpoints. Further, the incorporation of $var$ and $cov$ regularization terms is effective in maintaining diversity and removing redundancy. Through experimental results, we demonstrate that OTCXR identifies precise pathological regions relevant to the given task. These findings emphasize the importance of learning view-invariant features by achieving dense semantic invariance to align the representations in the SSL framework.

\bibliographystyle{ieee_fullname}
\bibliography{egbib}

\end{document}

\title{OTCXR: Rethinking Self-supervised Alignment using Optimal Transport for Chest X-ray Analysis}
\author{
Vandan Gorade\textsuperscript{1,\textdagger}, 
Azad Singh\textsuperscript{2,\textdagger}, 
Deepak Mishra\textsuperscript{2}\\
\textsuperscript{1}Northwestern University\\
\textsuperscript{2}Indian Institute of Technology Jodhpur\\
{\tt\small vandan.gorade@northwestern.edu, singh.63@iitj.ac.in,  dmishra@iitj.ac.in}\\
\textsuperscript{\textdagger}These authors contributed equally to this work.
}

\maketitle
%%%% ijcai24.tex
\section{Failure Case.}
Fig.~\ref{fig_cam:failure_cases} presents the failure case where all the baseline approaches and \new{OTCXR} failed, including the proposed one.

% \begin{figure*}[!ht]
% % \def\svgwidth{\columnwidth}
% \includegraphics[width=0.9\linewidth]{fig/IJCAI_rebuttal.pdf}
% % \includegraphics[width=0.5\columnwidth,scale=1.7]{abcd.png}
% % \vspace{-0.3cm}
% \caption{Diagnostic heatmaps for OTCXR and the baseline methods in addition to that in Figure 2 of the manuscript.}
% \label{fig_cam:failure_cases}
% % \vspace{-0.5cm}
% \end{figure*}

\begin{figure*}[!ht]
\centering
\includegraphics[width=0.9\linewidth,height=\linewidth,keepaspectratio,scale=0.8]{fig/IJCAI_rebuttal.pdf}
\caption{Diagnostic heatmaps for OTCXR and the baseline methods in addition to that in Figure 2 of the manuscript.}
\label{fig_cam:failure_cases}
\end{figure*}

\section{Visualization of the Transport Plan}
\begin{figure*}[t]
\centering
\def\svgwidth{\columnwidth}
\includegraphics[width=0.9\textwidth,scale=0.8]{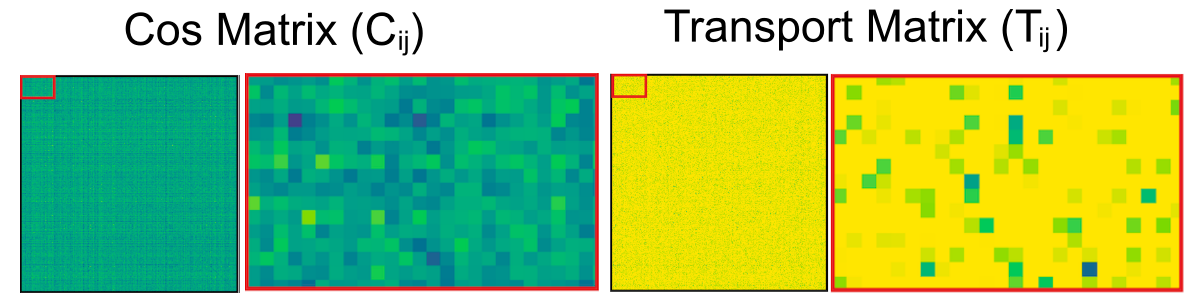}
\vspace{-0.2cm}
\caption{Aquired mean transport plan and cost matrix correspond to the samples in a batch after pre-training of 100 epochs.}
\label{fig_cam:OT_matrix}
% \vspace{-0.7cm}
\end{figure*}

Fig.~\ref{fig_cam:OT_matrix} presents the visualization of the transport
plan and the cost matrix for OTCXR.

\section{Motivation for the $R_s$ and $R_t$.}

The motivation for the $R_s$ and $R_t$ is to learn $\mu$ and $\nu$ over the visual feature space across diverse viewpoints. Without $R_s$ and $R_t$,  $\mu$ and $\nu$ are traditionally initialized as uniform distributions (L284-298), which treat all pixels equally, including irrelevant background pixels. However, using a cross-view attention mechanism, the proposed CV-SIM module (sec.3(2)) addresses this limitation by capturing intricate dependencies across different viewpoints. This enables dynamic focusing on discriminative pixels, ensuring that important regions 
are aligned effectively (sec.3(3)). 
% In other words, 
Discarding the CV-SIM module (Rs and Rt) would cause the model to over-represent irrelevant features(sec.4.1(Fig-2)), leading to poor performance (sec.4.1(Table-3)). 
% potentially overlooking crucial regions of interest in specific areas. 
% Further $var+cov$ terms introduce additional discrimination in the learned semantic features and leads to overall improvement 
% in terms of learning representation specific to ROIs.
% (sec.4.1(Table-3)). 
Further, Table 3 presents the results with uniform $\mu$ and $\nu$ (without CV-SIM module), and we observe a considerable degradation in the overall performance. Therefore, the CV-SIM module combined with OT is clinically relevant as it aids in capturing clinically significant information from medical images, enhancing diagnostic accuracy.\\

\section{Superiority over naively integrating OT in SSL framework.}
Simply integrating OT and contrastive learning involves aligning probits and logits simultaneously. In contrast, we propose an effective reformulation of the OT problem within the context
of SSL to achieve dense semantic invariance by introducing the novel CV-SIM module that utilizes a multi-head cross-view attention mechanism to extract subtle relationships and dependencies across different viewpoints, leading to the initialization of $\mu$ and $\nu$ distributions (Section 3(3)). Furthermore, unlike existing methods such as DenseCL, SimCLR, and PCRLv2, we shift the focus from traditional pixel-wise differences to dense feature maps, thereby capturing more meaningful semantic relationships and spatial information. Finally, the transport plan is computed using Sinkhorn's al-
algorithm, which is easily adaptable to batches of samples with
varying lengths, enabling GPU-friendly computations. Importantly, all of Sinkhorn’s operations are differentiable, optimizing the embedding with SGD, making it compatible with
deep learning frameworks, and reducing computational complexity. However, we agree that OT introduces some computational overhead, but this is during pre-training only. Meanwhile, there is no additional computational overhead in the downstream phase and for inference.

% \bibliographystyle{ieee_fullname}
% \bibliography{egbib}